# Collaborative Three-Tier Architecture Non-contact Respiratory Rate Monitoring using Target Tracking and False Peaks Eliminating Algorithms

Haimiao Mo, Shuai Ding*, *Member, IEEE*, Shanlin Yang, Athanasios V.Vasilakos, Xi Zheng, *Member, IEEE*

*Abstract*—**Monitoring the respiratory rate is crucial for helping us identify respiratory disorders. Devices for conventional respiratory monitoring are inconvenient and scarcely available. Recent research has demonstrated the ability of non-contact technologies, such as photoplethysmography and infrared thermography, to gather respiratory signals from the face and monitor breathing. However, the current non-contact respiratory monitoring techniques have poor accuracy because they are sensitive to environmental influences like lighting and motion artifacts. Furthermore, frequent contact between users and the cloud in real-world medical application settings might cause service request delays and potentially the loss of personal data. We proposed a non-contact respiratory rate monitoring system with a cooperative three-layer design to increase the precision of respiratory monitoring and decrease data transmission latency. To reduce data transmission and network latency, our three-tier architecture layer-by-layer decomposes the computing tasks of respiration monitoring. Moreover, we improved the accuracy of respiratory monitoring by designing a target tracking algorithm and an algorithm for eliminating false peaks to extract high-quality respiratory signals. By gathering the data and choosing several regions of interest on the face, we were able to extract the respiration signal and investigate how different regions affected the monitoring of respiration. The results of the experiment indicate that when the nasal region is used to extract the respiratory signal, it performs experimentally best. Our approach performs better than rival approaches while transferring fewer data.**

## I. INTRODUCTION

IN recent years, physical health monitoring has received a lot of attention [1]. One of the most crucial measures to assess physiological status is respiratory rate (RR) monitoring [2], particularly in cases where doctors are diagnosing respiratory conditions like chronic obstructive pulmonary disease (COPD), asthma, interstitial lung disease, pulmonary sarcoidosis, pneumoconiosis [3], and sleep apnea [4]. Traditional techniques typically use specialized medical equipment, such as a breathing belt or monitor, to monitor breathing [5], [6]. Due to the interaction of electrodes with skin, conventional RR monitoring approaches have been proved to be difficult and uncomfortable for users, which limits their use cases [7]. According to reports, environmental elements including lighting and motion artifacts have an impact on the photoplethysmography (PPG) signals that are acquired from visible facial videos to estimate respiration [8]. As a result, there is numerous research on RR monitoring using various techniques, including thermal infrared pictures [9], the Doppler effect [10], [6], optical flow field techniques [11], and breath sounds [12]. Infrared thermography can monitor human indicators in the dark and conduct large-scale population screening because it doesn't need a light source. This technology has a wide range of potential applications and an incalculable value for the future [13],[14]. It typically gathers respiratory signals from the neck arteries, nasal [15], and nostrils before estimating breathing.

Existing methods are difficult to correctly track the region of interest (ROI) of thermal infrared faces to obtain high-quality respiration signals. When compared to photographs taken with visible light, the contours of objects in infrared thermal images are blurrier.

The amount of interaction between users and the cloud is growing in real-world healthcare application situations [16], which causes delays in data transmission and service requests as well as compromising user security and privacy [17]. Patients' health and safety, as well as the decisions made by doctors, are directly impacted by data transmission delays, especially when those patients have been diagnosed with acute illnesses. Solutions based on deep learning have been popular in recent years across a variety of edge domains [18]. While considering the users' privacy and security, it efficiently resolves the latency of data transmission and service requests [19].

However, the technologies that are currently in use are unable to provide high precision and real-time RR monitoring for the demands of unique scenarios. We created a three-layer

This work was supported in part by the National Natural Science Foundation of China under Grant No. 91846107, in part by the Opening Foundation of the State Key Laboratory of Space Medicine Fundamentals and Application under Grant No. SMFA19K01.

Haimiao Mo, Shuai Ding, and Shanlin Yang are with the School of Management, Hefei University of Technology, Anhui Hefei 23009, China, also with the Key Laboratory of Process Optimization and Intelligent Decision-Making, Ministry of Education, China. And Yang is also with the National Engineering Laboratory of Big Data Distribution and Exchange Technologies, China (email: dingshuai@hfut.edu.cn, yangsl@hfut.edu.cn).

Athanasios V. Vasilakos is with the Center for AI Research (CAIR),University of Agder(UiA), Grimstad, Norway and with the College of Mathematics and Computer Science, Fuzhou University, Fuzhou 350116, China (Email: thanos.vasilakos@uia.no).

Xi Zheng is with the School of Computing, Macquarie University, Sydney, Australia(james.zheng@mq.edu.au).

architecture for non-contact RR monitoring (NRRM) to suit the demands of real-time medical situation detection and extract high-quality respiratory signals. We reduced the transferred data layer by layer based on the three-tier design. Additionally, it successfully prevents the transmission of patients' private information. By implementing an algorithm with eliminating false peaks in the edge layer, we increased the RR measurement's accuracy. Our main contributions are listed below.

*i)* We design a collaborative three-tier architecture to implement RR monitoring by task decomposition. Our architecture effectively safeguards patient privacy while reducing network burden. At the robot layer, the face background was removed by a face detection algorithm. Meanwhile, the tracking algorithm was used to locate the ROI from the infrared face. At the cloud layer, the face ROIs were used to extract the initial respiration signal. At the terminal layer, the physician could analyze the respiration signal to make an assisted decision.

*ii)* We ensure the quality of the respiration signal by designing a target tracking algorithm to locate the region of interest of the thermal infrared face. Meanwhile, we further improve the accuracy of respiration detection by designing an algorithm for eliminating false peaks.

*iii)* By collecting thermal infrared video from 15 volunteers (including medical staff, researchers, and patients in hospitals), we extract respiration signals from different regions of interest of the face and validate that the quality of respiration signals extracted from the nose or nostrils is better than other regions of the face.

The paper is organized as follows. Section II presents the related works of RR monitoring methods (including contact monitoring methods and non-contact methods) and deep learning on edges. Section III describes our three-tier architecture, including video capture and preprocessing at the robot layer, respiratory signal extraction at the cloud layer, respiratory signal processing, and decision support at the terminal layer. Section IV, Section V, and Section VI present our experiments, discussion, and conclusion, respectively.

## II. RELATED WORK

### A. Respiration Rate Monitoring Methods

#### 1) Contact Monitoring Methods

Human physiological parameters like RR, heart rate, and blood pressure can be effectively monitored using conventional methods. However, the majority of them are uncomfortable for the patient. Electrocardiograms (ECG) [20], pulse oximeters [21], and wearable technologies [22] are widely used contact-based physiological monitoring methods in healthcare institutions today [23]. These techniques necessitate skin-to-sensor direct contact [24]. They demand that patients wear specific monitoring equipment, such as thermistors [25], spirometers [26], and respiratory band sensors [5], while they are in a hospital for RR monitoring. These instruments typically only monitor one of the following factors: respiratory sounds [28], respiratory airflow, chest or abdominal breathing motions, or respiratory carbon dioxide emissions.

#### 2) Non-contact Monitoring Methods

Technologies that use visible light, the Doppler effect, and infrared thermal imaging are examples of non-contact monitoring techniques. Techniques for visible light typically incorporate remote PPG (rPPG). The primary component of rPPG approaches is the remote capture of facial footage by cameras to obtain PPG signals [7], [8]. It significantly affects how physiological measurements are estimated. Numerous methods for reducing noise have so far been suggested, including wavelet denoising, independent component analysis [28], and empirical mode decomposition.

According to the Doppler effect theory [29], RR is monitored by wifi, radar, and electromagnetic induction technology. The radar method uses bio-radar to transmit at wavelength into the human chest, which produces an echo signal as a result of the chest cavity's oscillating motion. Between the echo signal and the emitted signal, there is a phase difference, and the phase difference fluctuates with the displacement of the thoracic cavity. The respiratory signal can then be extracted from the echo signal for RR estimate. Respiratory function may alter the electrical conductivity of the lungs during routine physiological processes in the human body. Techniques for electromagnetic induction can be used to monitor RR by spotting changes in tissue conductivity [30].

With the emergence of a new generation of detectors, near and mid-infrared regions are also used for medical thermography [31]. Infrared thermography (IRT), often known as thermography, is a distant, non-contact monitoring technology that has gained popularity as a medical monitoring and diagnostic tool [32]. The major goal of thermography-based RR monitoring techniques is to detect respiratory signals by continually recording thermographic facial images and tracking ROIs (such as the carotid arteries and nose) [33]. To estimate RR and address apnea detection tasks, G. Scebba *et al.* suggested a novel approach based on multispectral data fusion [34]. Clinical investigations were conducted to determine whether infrared thermal imaging could be a clinically useful replacement for the RR monitoring techniques now used in neonatal care [35]. To extract the respiration signal, the aforementioned techniques, however, often track the region of interest of the human face in a scenario where the head is motionless or the human face may be recognized.

The device used for contact monitoring must come into direct contact with the human body to measure physiological markers. Prolonged contact can cause discomfort to the patient's body. Physiological health monitoring based on non-contact technology is non-intrusive and remotely managed. Other than IRT, there are non-contact monitoring techniques that are sensitive to outside influences like light, noise, and magnetic fields. In comparison to images taken with visible light, the contours of objects are somewhat fuzzy in infrared photographs. Present algorithms struggle to correctly locate the region of interest from faces to extract good respiration signals. In the event of face motion or occlusion, the present approaches are hardly ever able to meet the respiration signal extraction.

### B. Deep Learning on Edge

Massive volumes of data are produced by end devices like smartphones and Internet of Things (IoT) sensors, which are examined and processed using deep learning techniques to support decision-making. Deep learning inference and training, however, need a lot of processing power to operate quickly. Edge computing is a feasible method for placing a fine-grained grid of computing nodes close to terminal devices [36]. This method not only satisfies the high computational load and low latency requirements of deep learning for edge devices, but also offers further advantages including privacy, bandwidth efficiency, and scalability. Edge computing becomes increasingly potent as miniaturization progresses and processing power rises. Academia is increasingly interested in using edge computing to reduce stress on networks. By creating an edge computing method, we can lessen the stress on the network while accelerating system service response times. It opens the door for autonomous decision-making in the outskirts. Because the current edge nodes have a limited processing capacity, Li et al. developed a novel offloading strategy to optimize the performance of IoT deep learning applications through edge computing [18]. Zhang et al. developed a voting strategy that enables fog nodes to be selected as coordinators based on metrics of distance and processing capacity to speed up the training process [37]. By utilizing deep learning (DL) techniques in an edge computing environment, Chen et al. created a distributed intelligent video surveillance (DIVS) system to reduce the large network communication overhead [38]. Zhang et al. [39] presented a democratic, collaborative learning software that is built on fog. It used fog to resolve the data locality issue at each fog node and created a deep learning model that performed well in an IoT environment without a cloud. To address some of the issues with the conventional mobile cloud computing model, such as the unacceptably high system latency and short battery life of mobile devices, Liu et al. developed a food identification system based on edge computing [40].

Real-time performance needs to be higher in medical scenarios [17]. A high number of contacts between users and the cloud may cause communication latency. Even worse, it contributes to medical safety accidents and jeopardizes the patient's health diagnostic or life safety, especially in the case of those suffering from acute illnesses. It's critical to consider both the validity of physiological indicators and the impact of network congestion on real-time diagnosis in this circumstance Giving doctors decision support in real-world scenarios is only possible when both requirements are met.

### III. A COLLABORATIVE THREE-LAYER ARCHITECTURE FOR NON-CONTACT RESPIRATORY RATE MONITORING

### A. Overall Framework

Figure 1 depicts the collaborative three-tier architecture. A robot layer, cloud layer, and terminal layer are all parts of our architecture. Collaborative computing of three-tier architecture completes non-contact RR monitoring. i) At the robot layer, videos of faces from volunteers are captured by mobile robots. Additionally, a face detection technique is employed to identify key features of the face and acquire infrared face image sequences. Based on this, a tracking algorithm is made to find the ROIs from the infrared face, like the nose ROI. ii) ROI image sequences that do not contain faces are transferred to the cloud layer. By computing the average value of the pixels in each frame of the ROI image series, the respiratory signal of nose temperature variations is derived. iii) The terminal layer preprocesses the initial respiration signal, which is subsequently sent back to the doctor to help with decision-making. The signal preprocessing steps include detrending, normalizing, Butterworth filtering, and eliminating false peaks.

### B. Video Capture and Preprocessing at the Robot Layer

Mobile robots are deployed at the robot layer. The robot layer is mainly responsible for video capture and processing. As shown in Figure 1-(a), the video is captured by mobile robots. To find the key feature points of the face from the first frame, Google mediapipe [41], a common face detection technique, is employed. In this way, the coordinate information of the face (blue box) and face ROI (red box) is obtained, such as the nose. Due to the relatively blurry face outlines in thermal infrared images, a tracking algorithm is created to segment the nose region. If neither the face detection algorithm nor the tracking algorithm can obtain the region of interest of the segmented face, the information of the frame will not be transferred to the next layer. As a result, less time is wasted transmitting and processing pointless frames.

Accurate tracking of the face ROI is one of the key factors to obtain the respiration signal. To accurately extract spatiotemporal features and generate high-quality respiration signals, a Siamese network [42] is employed to track the facial ROI (such as the nose). Siamese network has two neural networks [43] (network 1 and network 2) with the same structure, and parameters can be shared between them. As a result, the Siamese network requires two inputs (input 1: $X$ and input 2: $Z$). The two feature extraction networks of the Siamese network map the two inputs to new spaces, $X'$ and $Z'$, as shown in Figure 2. By calculating the loss, the Siamese network assesses how similar the two inputs are to one another.

Our Siamese network includes two lightweight convolutional neural networks (CNNs). As demonstrated by the CNN design in Siamese network (Table 1), the CNN has five convolutional layers, four ReLU layers, and two Pooling layers. And the first to fourth convolutional layers (Conv-1~Conv-4) are followed by a ReLu layer. The relationship between convolutional layers' neighboring layers is described by the channel mapping characteristic between them. One CNN is used to extract the features of the searched image $Z$ and another CNN is also used to extract the features of the sample image $X$. The two CNN networks of the Siamese network extract the feature vectors of the two inputs respectively. The similarity of two vectors is measured by distance, such as Mahalanobis distance.

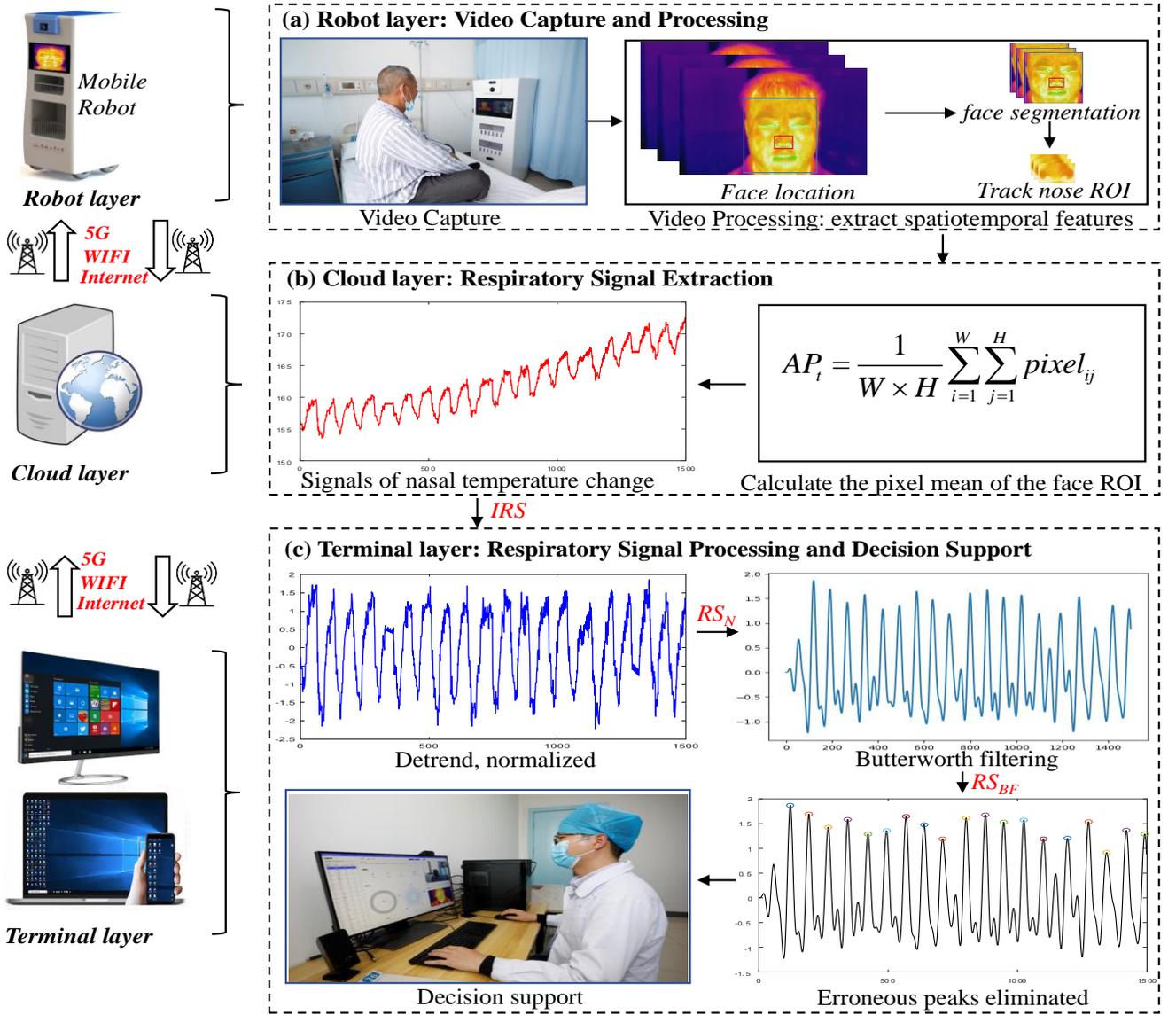

**Fig. 1.** The collaborative three-tier architecture for non-contact respiratory rate monitoring

According to the information of the target position of the previous frame, $N$ search boxes are generated in the current frame to mark candidate targets. The similarity between each candidate target and the target area of the previous frame is calculated through the Siamese network to form a similarity score map. According to the maximum value of the similarity score map, the target area of the current frame is determined.

In this case, $Z$ is an image containing an infrared human face. And $X$ is a target region of interest in image $Z$, such as the nose area. A sample image X chosen by the user serves as one input into the Siamese network, and a bigger search image Z serves as the other input. Simultaneously, the similarity between the search image Z and the sample image X is calculated by the computing resources from the robot layer to track the face ROI. A similarity score map is created from the robot layer by comparing the similarities between the sample image $X$ and various portions of the search image $Z$. The similarity score map is used to locate and track the target. The nose ROI is precisely found based on the largest value of similarity. A full convolutional network (FCN) with a single intercorrelation layer replaces the procedure of calculating the similarity score for each sector to lower the computational cost [44]. According to the knowledge of the maximum similarity, the target's position $roi_i = \left[ x_i^{min}, y_i^{min}, w_i, h_i \right]$ for the $i$-th frame is determined. The target's position is described by the coordinate information of a rectangle. The upper left corner of the rectangle that designates the $i$-th ROI area has coordinates of $(x_i^{min}, y_i^{min})$. The rectangle's width and height are denoted by $w_i, h_i$. The main steps of extracting the spatiotemporal features are described in Algorithm 1 and Figure 2. The face ROIs (such the nose ROI) sequence that do not include private information was sent to the terminal layer through the target tracking algorithm. By doing this, it is possible to significantly cut down on the resources needed for data transmission while yet protecting individual privacy.

**TABLE 1**
CNN DESIGN IN SIAMESE NETWORK

| Layer | In channels | Out channels | Kernel size | Stride |
|-------|-------------|--------------|-------------|--------|
| Conv-1 | 3 | 96 | 11 | 2 |
| MaxPool-1 | - | - | 3 | 2 |
| Conv-2 | 96 | 256 | 5 | 1 |
| MaxPool-2 | - | - | 3 | 2 |
| Conv-3 | 256 | 384 | 3 | 1 |
| Conv-4 | 384 | 384 | 3 | 1 |
| Conv-5 | 384 | 256 | 3 | 1 |

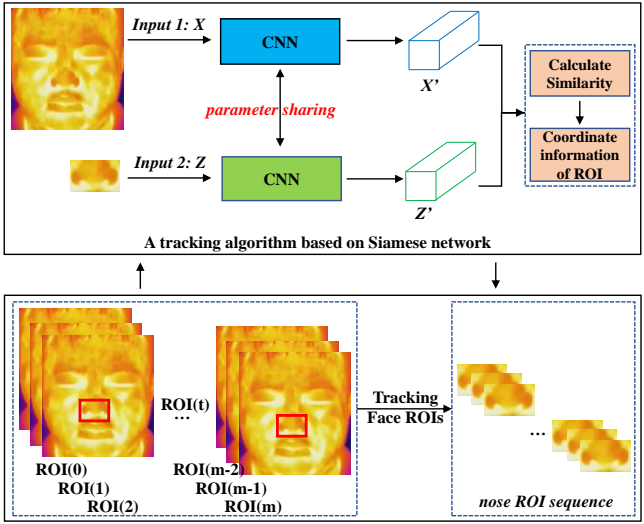

**Fig. 2.** Spatiotemporal features tracking.

### C. Respiratory Signal Extraction at the Cloud Layer

As shown in Figure 1, the robot layer receives image sequences of face ROIs (e.g. nose, nostrils) that do not contain private information. The initial respiration signal is extracted by processing the image sequence from the face ROIs and calculating the pixel average for each frame. Generally, there are three interconnected steps that make up the human breathing process: *i)* The exchange of gases between the lungs' alveoli and ambient air, as well as between those organs' alveoli and pulmonary capillary blood, is referred to as external breathing. *ii)* gas transportation in the blood. *iii)* The exchange of gases between blood and tissue cells is called internal respiration.

The majority of the gas exhaled while breathing is carbon dioxide, which is created by the body's metabolism when the nose exhales. The temperature of the exhaled gas is significantly higher than the temperature of the air around the nose. And the temperature of the inhaled air is significantly lower than the temperature of the exhaled gas of the nose. The air temperature around the nose changes slightly due to breathing, which is reflected in the thermal infrared image as a change in the pixel-average value.

As shown in Figure 1, the temperature signal fluctuation information of the ROI (such as nose ROI) is used to extract the initial respiration signal. After converting the face ROI to grayscale images [33], Equation 1 is used to determine the average pixels in each frame.

**Algorithm 1.** Extract spatiotemporal features of face ROI

**Input:** coordinate *of first frame face ROI and one-minute thermal infrared face video*

**Output:** *ROI={roi₁, roi₂, ..., roiₘ}*

The algorithm steps are as follows.

**Step 1:** *Process a minute of the video frame by frame to get the sequence of images in the video;*

**Step 2:** *Use a face detection algorithm to locate the face key points of the first frame, and mark the face ROI coordinate information. The first frame image and its face key point information are used as the initial input of the tracking algorithm;*

**Step 3:** *For the i-th frame, the similarity between $X_i$ and $Z_i$ is calculated, and the i-th position information $roi_i$ of the face ROI is determined by the tracking algorithm;*

**Step 4:** *Until all image sequences are processed, and finally output the location information set of the image sequence ROI={roi₁, roi₂, ..., roiₘ}.*

$$AP_t = \frac{1}{W \times H} \sum_{i=1}^{W} \sum_{j=1}^{H} pixel_{ij} \qquad (1)$$

Where $AP_t$ is the pixel-average value of the image at frame *t*. And $pixel_{ij}$ is the pixel value with coordinates $(x_i, y_i)$. The signal produced by averaging the ROI pixels across each frame of the image stream is known as the initial respiration signal (IRS). Moreover, the IRS is described as follows.

$$IRS = \{AP_1, AP_2, ..., AP_m\} \qquad (2)$$

Where *m* is the total number of frames of a thermal infrared video.

### D. Respiratory Signal Processing and Decision Support at Terminal Layer

Since the IRS contains a lot of noise, it needs to be preprocessed at the terminal layer, which includes detrending, normalization, Butterworth filtering, and eliminating false peaks. In addition, the IRS and other signals generated by the preprocessing process are called respiration signals.

#### 1) Detrending

Due to oscillations and low-frequency components in the IRS, the respiration signal's quality may be impacted. The instability of the signal acquisition equipment and the sensitivity to environmental disturbances often cause the signal to deviate from the baseline over time. The entire process of deviating from the baseline over time is called the "trend term". It influences the effectiveness of respiratory signal extraction and the precision of RR monitoring. A previous method of smoothing eliminates the trend of IRS [45]. The IRS is divided into segments and a least-squares fit of the straight line is computed for each segment [46]. The trend term function is the fitted straight line of each signal. The relevant trend term function value is subtracted from the IRS to obtain the respiration signal without a trend term. The signal obtained by IRS after detrending processing is called the detrended respiration signal ($RS_D$).

#### 2) Normalization

Data can be normalized in numerous ways. Additionally, they are divided into three categories: linear methods (such as the extreme value method and the standard deviation method), folding methods, and curvilinear methods (e.g., half-normal distribution). The $RS_D$ is normalized using the procedure listed below [45].

$$RS_N = \frac{RS_D - \mu}{\delta} \tag{3}$$

Where $\delta$ is a standard deviation, $\mu$ is the mean of the $RS_D$. And the normalized $RS_D$ is called $RS_N$ (as shown in Figure 1).

### 3) Butterworth Filtering

Frequency domain characteristics are frequently employed in non-contact physiological monitoring techniques to collect physiological signs. The frequency-domain signals connected to physiological indicators are often obtained using band-pass filters in this technique. The frequency range of normal breathing is 0.15 to 0.40 Hz or 0.15 to 0.7 Hz. Transform a time domain signal to a frequency domain signal using a Fast Fourier Transform (FFT). The time-domain signal is transformed into a frequency-domain signal using the FFT. The respiratory signal in the normal range is preserved by the band-pass filtering method, while the remaining frequency ranges are set to zero. By doing this, the respiration signal's noise is removed.

In addition, frequency-domain data important for RR analysis should be extracted. A more consistent respiratory signal is associated with the low frequency. The high-frequency portion is filtered out while the low-frequency portion is kept. The Butterworth filter has a maximum flat amplitude characteristic in the passband, and the positive frequency amplitude monotonically drops with increasing frequency [47]. It is typically applied as a low-pass filter. Finally, the Butterworth filter eliminates noise from the $RS_N$. As shown in Figure 1, the signal generated by $RS_D$ after Butterworth filtering is called $RS_{BF}$.

### 4) Eliminate False Peaks

As shown in Figure 1, the $RS_{BF}$ acquired after preprocessing the IRS can still contain certain peaks that were incorrectly noticed for a variety of reasons. It is necessary to remove these false peaks from $RS_N$. We set an amplitude threshold to eliminate false peaks. The main steps for eliminating false peaks are the following.

*i)* The low-frequency signal will have an impact on the error peaks, which will differ from the true respiration signal. These waveforms' peaks are less than zero, hence they are disregarded to produce candidate waveforms.

*ii)* Calculate the average value of the candidate wave amplitude $awa$.

*iii)* Calculate the peaks threshold $thr$ according to the following formula 4. If the amplitude value of candidate peaks is greater than or equal to $thr$, one breath is accumulated.

$$thr = awa \times (1 - \xi) \tag{4}$$

The number of peaks obtained after eliminating false peaks is called $PN$. Then we mark the index of the first peak and the last peak as $FP$ and $LP$, respectively. The average distance between two adjacent peaks is $ADP$. And the total number of frames of thermal infrared video acquired in one minute is $TF$. The final respiration rate is calculated as follows [48].

$$RR = \frac{FP + (TF - LP)}{ADP} + PN \tag{5}$$

Since the information of $[0, FP]$ and $[LP, TF]$ is ignored in the process of calculating the RR, the number of breaths in this part needs to be considered.

## IV. EXPERIMENTS

### A. Experimental Setup

At the Second Affiliated Hospital of Anhui Medical University, medical and nursing staff were involved in collecting clinical data. Fifteen healthy participants were recruited. Their age range was 20 to 31 years old, and their weight range was 46 to 105 kg. All participants voluntarily participated in our research and signed an informed consent form. The local ethics committee approved the study. We used a mobile robot with an infrared camera (Guide Sensmart IPT640) to collect facial video. The wavelength of the camera is 8 to 14. The camera's temperature measurement range is -20 to 150 degrees Celsius, and the working environment temperature is -10 to 50 degrees Celsius. The resolution of the infrared camera is 640 × 480 pixels, and the noise equivalent temperature difference (NETD) is less than or equal to 60 $MK$. The parameters of the camera lens are 30°×23°/20 $mm$. We collected data in an environment with a temperature of 23~26°C. We started collecting data on March 14, 2020. During the data collection process, volunteers were asked to sit in front of the mobile robot with their heads allowed to move freely. The amount of data collected was between 5 and 30 minutes per day. All participants were allowed to perform unconscious head movements during the data collection process. As of April 14, we collected a total of 305 minutes of data. Some of our data were invalid due to human factors and equipment failure. After removing the invalid data, we used 274 minutes of useful video for our experiment.

Firstly, we preprocessed the collected video at the robot layer. We also created a comparative experiment to test the efficacy of our three-tier architecture in reducing network load. In other words, in the robot layer, we segmented various ROI of facial image sequences before transferring them to the following layer.

By choosing several ROIs from the infrared face, we then looked into how they affected the precision of RR monitoring. To assess the effectiveness of various comparison techniques, we employed mean absolute error (MAE) and root mean square error (RMSE).

$$MAE(X, h) = \frac{1}{m} \sum_{i=1}^{m} \left| h(x^{(i)}) - y^{(i)} \right| \tag{6}$$

$$RMSE(X, h) = \sqrt{\frac{1}{m} \sum_{i=1}^{m} (h(x^{(i)}) - y^{(i)})^2} \tag{7}$$

Where $h(x^{(i)})$ is the ground truth of the RR in the *i*-th video, $y^{(i)}$ is the predicted value of the corresponding RR, and $m$ is the total number of respiration data.

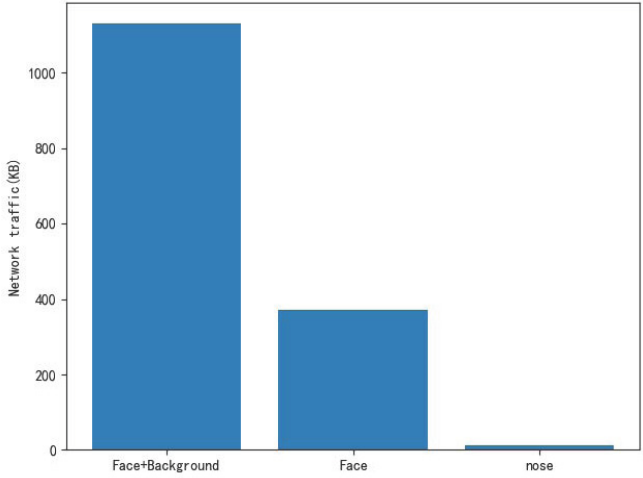

**Fig. 3.** Comparison of network traffic with different ROI.

Thirdly, the performance of our method was evaluated by comparative experiments with other state-of-the-art methods. The comparison methods include NRRM without an error peak elimination (NRRM-EEP), Gaussian window (GW) [8], and frequency domain analysis method (FDAM) [49]. And NRRM-EEP is an NRRM method without eliminating false peaks. GW is a fixed-width Gaussian windowing scheme for respiration signals after Butterworth filtering. FDAM is a method that converts the time domain signal to the frequency domain to calculate the RR. And our codes are available at *https://github.com/1286705494/Non-contact-Respiratory-Rate-Monitoring*.

Finally, we evaluated the consistency between the RR monitored by different methods and the true RR. The Bland-Altman (BA) method is usually used to evaluate the consistency between the two methods [34]. The problem of comparing two testing methods is often faced in clinical studies. For example, *A* is a classical method and *B* is a new method. It is limited by the 95% agreement between the measurements of the two methods. The mean of the true and predicted values is the horizontal axis. We drew a scatter plot and marked the 95% limits of agreement. The maximum error allowed in clinical practice is combined to conclude whether the two methods are consistent or not.

### B. Cost and Time Comparison Experiment

Images with varying sizes were chosen to evaluate the effects on network traffic, including the original image (background + face), face and nose area (ours). Furthermore, we evaluated whether our model could achieve real-time performance in terms of communication cost and computational cost when the nose was used as the region for extracting respiration signals. The Intel (R) Core (TM) i11-11800H, clocked at 2.3 GHz, the NVIDIA Geforce RTX3050, and 16 GB of RAM make up the robot layer's configuration. A workstation equipped with an Intel (R) Core (TM) i7-10510U processor with a maximum turbo frequency of 4.80 GHz and 16.0 GB of RAM was used to implement the cloud layer. A notebook was used to implement the terminal layer. It has an Intel (R) Core (TM) i7-8565 processor with a base clock of 1.8 GHz and a maximum turbo frequency of 4.6 GHz, an NVIDIA Geforce MX150 graphics card, and 8GB of RAM.

#### 1) Comparison of network traffic with different ROIs

We designed experiments to compare the impact of the original image (background + face), face, and nose area (ours) on network traffic. First of all, we transferred the original image sequence without any preprocessing from the robot layer to the cloud layer and used it to extract the IRS. Secondly, we segmented the original picture at the robot layer and then transferred them to the cloud layer to extract respiratory features. In the last case, we segmented the face background at the robot layer. And the nose was located and segmented at the robot layer, and then the nose region was transmitted to the cloud layer to acquire respiration signals.

The experimental results, which are depicted in Figure 3, demonstrate that segmenting the original image and subsequently locating the nose region can significantly lessen network traffic. To reduce network traffic by almost two times, the face is segmented before being sent to the cloud layer. Furthermore, the robot layer segments the nose region, which is relayed to the cloud layer after, greatly reducing network demand. It is reinforced by the fact that compared to transmitting the original image, our three-layer architecture may efficiently cut network traffic by more than 100 times.

#### 2) Communication and Computational Costs

**TABLE 2**
COMMUNICATION AND *COMPUTATIONAL* COSTS (SECONDS)

| TIME COSTS | MIN | MAX | MEAN | STD |
|---|---|---|---|---|
| COMM. COST | 23.40 | 27.19 | 23.50 | 0.24 |
| COMP. COST | 24.91 | 26.32 | 25.30 | 0.28 |

Since our aim is to meet real-time constraint, we evaluated mainly on the communication latency and computation processing latency which are vital for our target applications. The CNNs used are relatively lightweight and computation cost in terms of cpu and memory usage, compared with our hardware settings, are trivial. As shown in Table 2, the experimental results of the communication cost and computational cost of our model in 274 videos are presented. Each video lasts for roughly one minute. Our model makes it possible to monitor RR in real-time. Because the sum of the communication cost and computational cost of our model does not exceed one minute when processing one minute of video. In terms of communication cost, our model's minimum, maximum, mean, and standard deviation are 23.40, 27.19, 23.50, and 0.24 seconds, respectively. In terms of computational cost, the minimum, maximum, mean, and standard deviation of our model are 24.91, 26.32, 25.30, and 0.28 seconds, respectively.

### C. Effect of different ROI for extracting the respiratory signal

As shown in Figure 4, we defined ROIs of the nose (ROI-1),

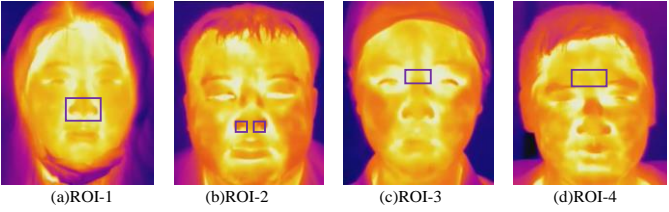

**Fig. 4.** Motion Non-sensitive ROIs. (a)ROI-1: nose, (b)ROI-2: nostrils, (c)ROI-3: nasal root, (d) ROI-4: the area above the nasal root.

nostrils (ROI-2), nasal root (ROI-3), and the area above the nasal root (ROI-4) from the infrared face as motion-insensitive regions since they are not easily affected when speaking [50]. The temperature fluctuations in ROI-1 and ROI-2 are more obvious than in other areas when people breathe regularly [8], [51]. We chose ROI-3 and ROI-4 with rich blood vessels based on the facial blood vessel distribution [52]. As shown in Figure 5, we chose a wider nose (ROI-5, marked by a yellow box), half-face (ROI-6, marked by a green box), and face (ROI-7, marked by a blue box) to extract respiratory signals and compared them with ROI-1 and ROI-2 to investigate the impact of the ROI's size on the results of the respiratory monitoring.

### 1) Motion Non-sensitive ROIs

We implemented our scheme to evaluate the effectiveness by tracking the motion non-sensitive ROIs (shown in Figure 4) to extract different respiratory signals and calculate the RR. According to Equation 4, we set $\xi$ to be 0.20, 0.25, 0.30, 0.35, and 0.40, respectively, and the experimental results are shown in Table 3 and Figure 6. According to the results in Table 3, the results of the respiration rate extracting respiratory signals with ROI-1 are better than the others when the amplitude threshold is 0.25. At this time, the overall MAE and RMSE are 0.7952 and 1.1169, respectively. Moreover, the experimental results obtained by ROI-3 and ROI-4 were poor.

The BA method is used to evaluate the consistency between RR monitoring by NRRM and the ground truth of RR. The mean of the ground truth and predicted values is the $x$-axis. The difference between measurement results is the $y$-axis. The abscissa is the average value of the measurement data of the two methods, and the ordinate is the measurement difference between the two methods. The BA chart has three lines in total. The center line shows the average difference value, while the top and lower lines show the maximum and lower limits of the 95 percent consistency limit (ie, 1.96 standard deviations).

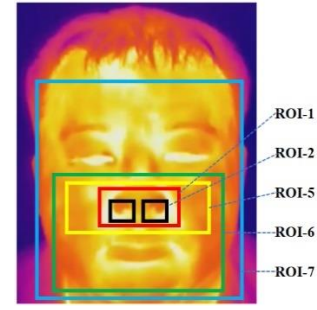

**Fig. 5.** ROIs with different sizes. (ROI-5: a wider nose, ROI-6: half-face, ROI-7: face, ROI-2 < ROI-1 < ROI-5 < ROI-6 < ROI-7)

We drew the scatter plot and marked the 95% consistency limit. Most of the scattered points in Figure 6-(a) occur within the range [-2.0317, 2.3120] of the upper and lower limits, which is quite narrow compared to other BA plots in Fig. 6. The consistency between the upper and lower bounds in Figure 6-(b) falls within the range [-2.7738, 2.8346]. For ROI-3 and ROI-4, there is no discernible difference between the top and lower consistency bounds. In using ROI-1 to extract respiration signals, ROI-1 performs the best when compared to other regions of interest from faces. In other words, the consistency analysis of the RR derived by ROI-1 is the best when $\xi$ is set to 0.25.

### 2) ROIs with Different Sizes

As shown in Figure 5, the order of the size of the ROI area is ROI-2 < ROI-1 < ROI-5 < ROI-6 < ROI-7. To investigate the effect of ROI size on respiration monitoring, we selected ROIs of varied sizes. Table 4 displays their contrasted experimental results. And ROI-1 produced superior experimental findings than ROI-2 when ROI-1 produced better experimental results than ROI-2 when estimating respiratory monitoring. However, the motion-sensitive area will be included due to the expansion of the chosen ROI area. Motion-sensitive areas include the mouth, eyes, and areas of facial muscle movement due to speaking. As a result, the respiration signal may become noisy and the accuracy of respiration monitoring may be impacted. In short, the experimental findings indicate that selecting a bigger ROI region does not necessarily result in a higher quality respiratory signal extraction.

**TABLE 3**
Total MAE and RMSE for Tracking Different ROIs

| ROI | MAE | | | | | RMSE | | | | |
|---|---|---|---|---|---|---|---|---|---|---|
| | $\xi = 0.2$ | $\xi = 0.25$ | $\xi = 0.3$ | $\xi = 0.35$ | $\xi = 0.4$ | $\xi = 0.2$ | $\xi = 0.25$ | $\xi = 0.3$ | $\xi = 0.35$ | $\xi = 0.4$ |
| ROI-1 | 0.9073 | **0.7952** | 0.8400 | 1.1284 | 1.6025 | 1.3053 | **1.1169** | 1.2484 | 1.5723 | 2.1456 |
| ROI-2 | 0.9749 | 0.9342 | 0.9842 | 1.2810 | 1.646 | 1.5158 | 1.4214 | 1.4518 | 1.8533 | 2.2405 |
| ROI-3 | 3.5934 | 4.0160 | 4.6446 | 5.4512 | 6.3448 | 4.4682 | 4.9190 | 5.5790 | 6.2742 | 7.1454 |
| ROI-4 | 3.5510 | 4.0472 | 4.6951 | 5.4995 | 6.2748 | 4.4024 | 4.9182 | 5.5903 | 6.4033 | 7.1747 |

**TABLE 4**
Effects of different sizes of ROI on respiratory monitoring results

| ROI | MAE | | | | | RMSE | | | | |
|---|---|---|---|---|---|---|---|---|---|---|
| | $\xi = 0.2$ | $\xi = 0.25$ | $\xi = 0.3$ | $\xi = 0.35$ | $\xi = 0.4$ | $\xi = 0.2$ | $\xi = 0.25$ | $\xi = 0.3$ | $\xi = 0.35$ | $\xi = 0.4$ |
| ROI-2 | 0.9749 | 0.9342 | 0.9842 | 1.2810 | 1.646 | 1.5158 | 1.4214 | 1.4518 | 1.8533 | 2.2405 |
| ROI-1 | 0.9073 | **0.7952** | 0.8400 | 1.1284 | 1.6025 | 1.3053 | **1.1169** | 1.2484 | 1.5723 | 2.1456 |
| ROI-5 | 7.2433 | 8.3300 | 9.4774 | 10.7084 | 11.8251 | 8.1113 | 9.0776 | 10.2012 | 11.3390 | 12.3937 |
| ROI-6 | 6.4604 | 7.5083 | 8.6124 | 9.6921 | 11.0807 | 7.4742 | 8.3853 | 9.4289 | 10.3862 | 11.7340 |
| ROI-7 | 7.2218 | 8.2712 | 9.5406 | 10.5566 | 11.9750 | 8.3243 | 9.3165 | 10.5018 | 11.4696 | 12.7533 |

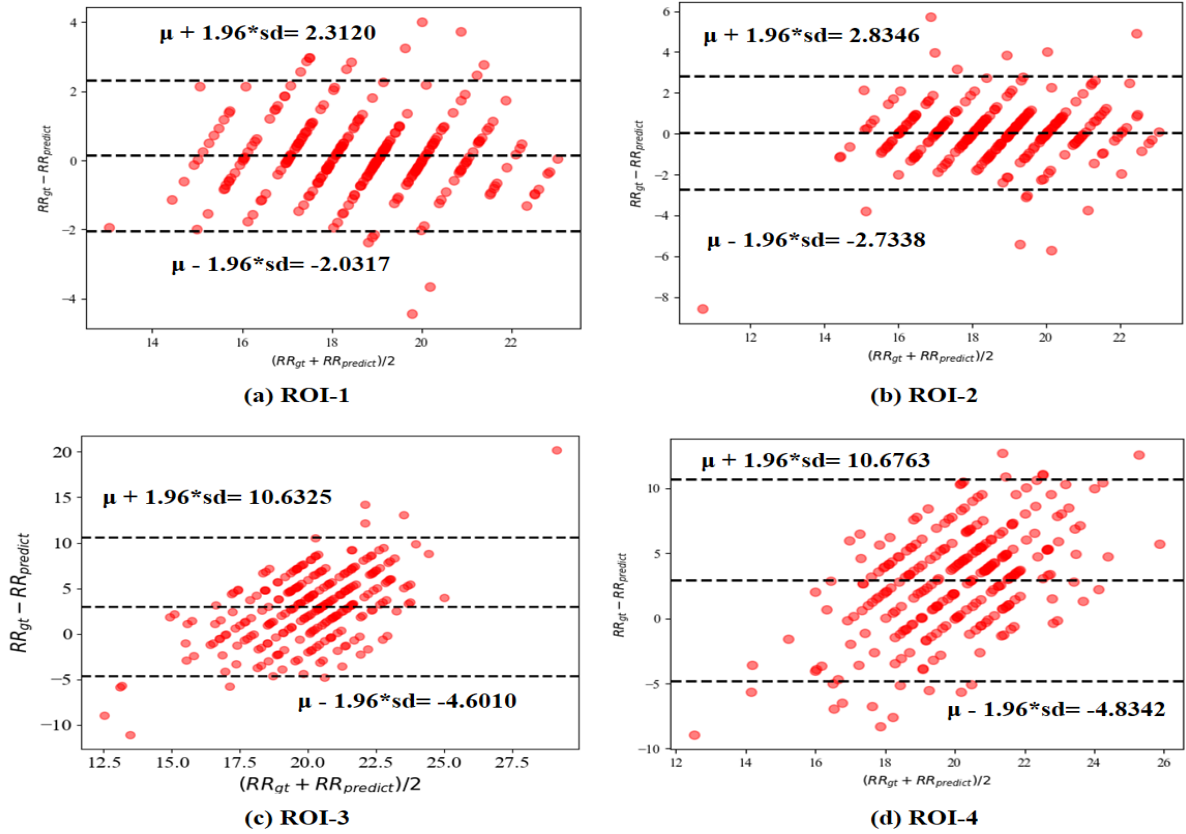

**Fig. 6.** The comparison of Bland-Altman plots for four ROIs ( $\xi$ =0.25).

**TABLE 5**
TOTAL MAE AND RMSE OF GW WITH DIFFERENT WIDTH

| Evaluation | w=50 | w=60 | w=70 | w=80 | w=90 | w=100 | w=110 | w=120 | w=130 | w=140 | w=150 |
|---|---|---|---|---|---|---|---|---|---|---|---|
| MAE | 8.9772 | 6.4491 | 4.4591 | 2.9535 | 1.9057 | 1.2824 | 1.0015 | 0.8704 | **0.8359** | 0.8791 | 0.9661 |
| RMSE | 9.9065 | 7.4396 | 5.4025 | 3.8177 | 2.5631 | 1.7995 | 1.4165 | 1.2548 | **1.1734** | 1.2227 | 1.4308 |

### D. Comparison with Other RR Monitoring Methods

To extract the RR signal, we decided to use nasal ROI. As a result, comparisons were done with other RR monitoring techniques, such as GW [8], FDAM [49], and NRRM without elimination of error peaks (NRRM-EEP).

To process the respiratory signals, the Gaussian window's width is adjusted to 50, 60, 70, 80, 90, 100, 110, 120, 130, 140, and 150. Table 5 presents the experimental outcomes. The approach operates most effectively when the width of the

**TABLE 6**
THE COMPARISON OF TOTAL MAE AND RMSE FOR FOUR METHODS

| Eva. | FDAM[49] | Ours-EEP | GW [8] | Ours |
|---|---|---|---|---|
| MAE | 6.8131 | 15.4377 | 0.8359 | **0.7952** |
| RMSE | 10.0692 | 15.8026 | 1.1734 | **1.1169** |

Gaussian window is 130. The MAE=0.8359 breaths/min (bpm) and RMSE=1.1734 bpm corroborate it.

The comparison of MAE and RMSE for the four approaches is summarized in Table 6. The best result is shown by our approach. Our method demonstrates a reduction in MAE and RMSE of 5.12 and 5.06 percent, respectively, when compared to the second-ranked GW method. The lowest MAE=0.7952 bpm and RMSE=1.1169 bpm corroborate it. With an MAE of 0.8359 bpm and an RMSE of 1.1734 bpm, GW comes in second. FDAM ranks third. Additionally, the RMSE=10.0692 bpm and MAE=6.8131 bpm support it. The lowest performance is displayed by NRRM-EEP (also called ours-EEP). It is supported by MAE=15.4377 bpm and RMSE=15.8026 bpm.

The upper and lower bounds of Figure 7-(a)'s consistency are relatively tiny in comparison to the other BA plots in Figure 7, and the majority of the scatter lies within their respective ranges. The consistency of Figure 7-(c) is closely comparable, and several of its scatters are outside the range [-4, 4]. The areas that are enclosed by the upper and lower consistency constraints for the other two baseline techniques are quite substantial. Their effectiveness is far lower than our approach. As a result, the performance of our method (NRRM) is the best. It is backed by the smallest interval and the lowest mean of difference. Compared with the other three methods, $\mu$ of NRRM is reduced by 0.2330 bpm, 6.6089 bpm, and 15.2975 bpm, respectively.

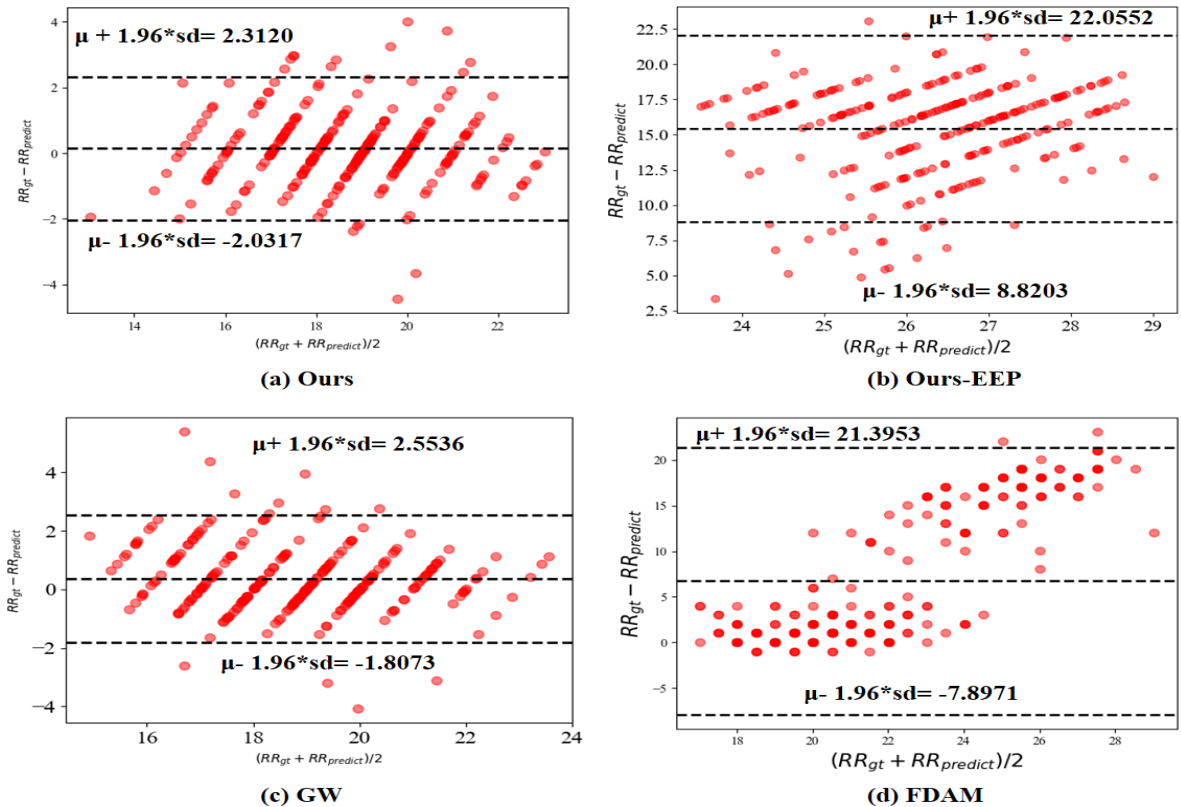

**Fig. 7.** The comparison of Bland-Altman plots for four methods.

## V. DISCUSSION

Edge computing is an important technology for IoT services [53], [54]. The vast amount of data gathered from IoT devices is difficult for centralized cloud computing organizations to process and analyze because of the limited network performance for data transfer [55]. The preprocessing procedure greatly decreases the transferred data as edge computing moves the computational activities from the centralized cloud to the edge close to the IoT devices. We decomposed the RR computing task layer by layer with a collaborative three-tier architecture. At the robot layer, we achieved face background segmentation and eliminated redundant information. At the same time, we located ROIs (such as the nose) in the segmented faces. The size of the nose region used to extract breath feature data is much smaller compared to the original image data. At the cloud layer, ROI image sequences from the robot layer are used to extract the respiration signal. At the terminal layer, we processed the respiration signal from the cloud layer. Our method efficiently reduces network load while maintaining volunteer privacy.

Due to the characteristics of thermal infrared technology, RR can be monitored in low light conditions even in dark environments. However, the thermal infrared camera's blurry image of the face makes it difficult for conventional techniques to segment the region of interest in the face. It affects the effectiveness of respiratory signal extraction, which in turn reduces the accuracy of RR monitoring. The NRRM not only overcomes the problems of human pain and annoyance brought on by conventional wearable monitoring technology, but it can also detect respiratory rate generally under the circumstance of head movement and facial occlusion. The quality of respiration signal from ROI-1 is superior to that of other ROIs. The air near the nose experiences large temperature variations during human breathing, which makes it easier for a thermal infrared camera to record these changes. This is one of the key explanations for why the ROI-1 performs the best when used to extract respiration signals. Only temperature differences of plus or minus 0.5°C can be detected by the camera itself. Although there are more blood vessels in the ROI-3 and ROI-4 regions, it is impossible to capture the subtle changes in temperature due to respiration. It is the primary cause of ROI-3 and ROI-4's subpar effectiveness in extracting respiratory signals. Furthermore, when the threshold value of the amplitude is set to 0.25, the ROI-1's RR outperforms the others.

We also chose ROIs of various sizes to extract respiration signals and looked into how these affected the outcomes of respiration monitoring. As the ROI region increases, the respiration signal becomes noisier due to the effect of motion-sensitive locations (such as the mouth and eyes). In other words, eye movements, lip motions during a speech, and facial muscle movements show up as typical pixel changes in the image, which in turn causes the respiration signal to become noisier. As a result, the size of the ROI has no impact on the effect of respiration monitoring when the chosen ROI does not contain motion-sensitive regions. In future work, we will collect infrared video data of faces in different scenes, enhance the quality of respiration signals and improve the robustness of respiration evaluation through the complementarity of ROIs.

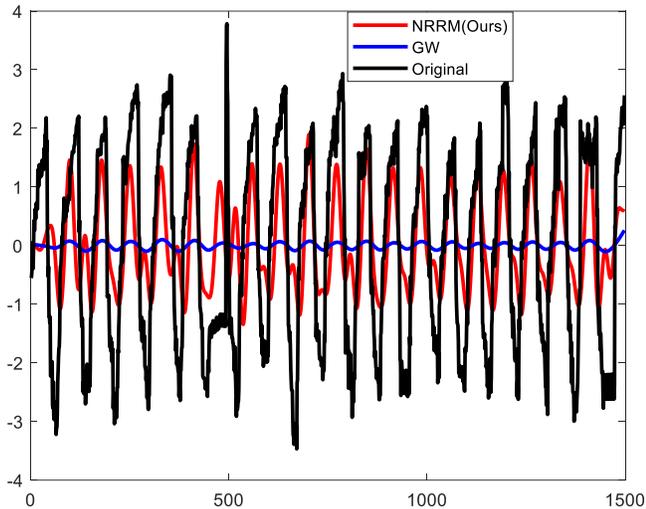

**Fig. 8.** The plots of NRRM(Ours), GW, and the original signal

The respiratory signal is processed by GW with a fixed width following Butterworth filtering in the comparison experiments. The GW method uses a fixed window size to process the respiration signal, which may miss some peaks. As the instantaneous respiration rate changes, the peak distance changes as well. Unintentional head motions cause motion artifacts, which can contaminate the respiratory signal and impair the FDAM's accuracy. NRRM-EEP calculates RR without eliminating false peaks, and it counts more peaks that should not be retained. The experimental results shown in Table 6 and Figure 7 demonstrate that NRRM successfully overcomes the shortcomings of the existing methods described above by setting amplitude thresholds.

The plots of GW, NRRM, and the original signal are displayed in Figure 8. Both GW and NRRM use the same signal preprocessing techniques such as detrending, normalization, and Butterworth filtering. However, the steps applied to eliminate false peaks are different. The former removes erroneous peaks by establishing an amplitude peak threshold, whereas the latter employs a polynomial Gaussian fitting approach. And the experimental outcomes between them did not differ significantly. Furthermore, it is evident from Figure 8 that NRRM and the original signal have a greater correlation of preserved peaks than GW with the original signal. The peak amplitude and RR from the respiration signals are used by medical professionals to aid in the identification of specific diseases [56],[57]. The correlation between signals is measure by the respiration amplitude, and the accuracy of the respiration rate. As seen in Fig. 8, the signal extracted using our method is better able to recover the respiration amplitude of the original signal when compared to the GW method. Furthermore, MAE and RMSE are used to evaluate the accuracy of respiration rate. Our method extracts the signal to estimate the respiration rate, and the corresponding MAE and RMSE perform better than the GW method. Therefore, our proposed method is superior to GW.

Prospects for widespread use are abundant in the medical industry. Due to the limited amount of data, we need to collect more data and explore related future work issues. Furthermore, to establish a solid scientific foundation for the implementation of comprehensive health monitoring, we also need to incorporate more contact and non-contact physiological data into our system.

## VI. CONCLUSION

In this paper, we proposed a collaborative three-layer architecture. Our three-layer architecture decomposes RR monitoring tasks layer by layer through the robot layer, cloud layer, and terminal layer. It efficiently preserves individual privacy while simultaneously reducing data transfer. Moreover, we designed a target tracking algorithm and an algorithm to eliminate false peaks to ensure the quality of respiratory signal extraction, thereby improving the accuracy of RR monitoring. Finally, we collected data sets from clinical scenarios and verified that the quality of the respiratory signal from the nose or nostril region was the best. The MAE and RMSE of the NRRM proposed in this research are at least 5.12 percent and 5.06 percent less accurate than the competition's current approaches. Our method's consistency assessment is superior to other comparison techniques. In other words, our NRRM not only determines RR in real-time but also makes sure that RR monitoring is accurate. In this situation, real-time effective respiratory signals can provide information to enhance the medical staff's decision-making if more physiological data is incorporated into the system. Furthermore, numerous scenarios in future unmanned hospitals can utilize our three-tier design. More non-contact and wearable physiological monitoring data, such as blood pressure, heart rate, heart rate variability, and facial temperature, will be incorporated in upcoming studies. We'll also look more closely at the relationship between physiology and psychology. In the future, our technique will be used in situations like long-term crew health protection while sailing, hospital ward rounds, public health, and aerospace by incorporating the understanding of mental health auxiliary treatment.

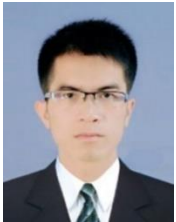

**Haimiao Mo** is a Ph.D. student at the School of Management, Hefei University of Technology, China. He is also a visiting Ph.D. student at the School of Computer Science and Engineering, Nanyang Technological University funded by China Scholarship Council.

His main research interests include noncontact health monitoring, healthcare Management, data mining, and machine learning.

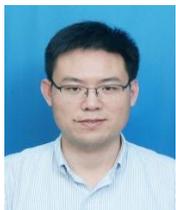

**Shuai Ding** (Member, IEEE) is a professor of information systems at the School of Management, Hefei University of Technology, China. He received his Ph.D. in MIS from the Hefei University of Technology in 2011. He has been visiting the University of Pittsburgh. His research interests include social networks, information artificial intelligence, cloud computing, and business intelligence. He has published more than 40 high-quality publications in international journals (IEEE Transactions on Knowledge and Data Engineering, ACM Transactions on Knowledge Discovery from Data, ACM Transactions on Internet Technology, Decision Support Systems, IEEE Journal of Biomedical and Health Informatics).

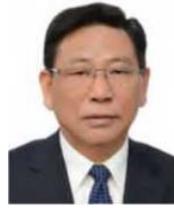

**Shanlin Yang** is a member of the Chinese Academy of Engineering and the leading Professor in management science and information system at the School of Management, Hefei University of Technology. He is the director of the academic board of Hefei University of Technology, and the director of the National-Local Joint Engineering Research Center of Intelligent Decision and Information System. He has won 2-second class prizes for State Scientific and Technological Progress Award, and 6 first-class prizes for provincial and ministerial level science and technology awards. His research interests include information systems, social networks, cloud computing, and artificial intelligence.

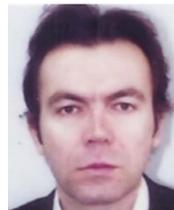

**Athanasios V. Vasilakos** is with the Center for AI Research (CAIR), University of Agder (UiA), Grimstad, Norway, and with the College of Mathematics and Computer Science, Fuzhou University, Fuzhou 350116, China. He served or is serving as an Editor for many technical journals, such as the IEEE TRANSACTIONS ON NETWORK AND SERVICE MANAGEMENT, IEEE TRANSACTIONS ON CLOUD COMPUTING, IEEE TRANSACTIONS ON INFORMATION FORENSICS AND SECURITY, IEEE TRANSACTIONS ON CYBERNETICS, IEEE TRANSACTIONS ON NANOBIOSCIENCE, IEEE TRANSACTIONS ON INFORMATION TECHNOLOGY IN BIOMEDICINE, ACM Transactions on Autonomous and Adaptive Systems, the IEEE JOURNAL ON SELECTED AREAS IN COMMUNICATIONS. He is WoS highly cited researcher (HC) from 2016 to 2021.

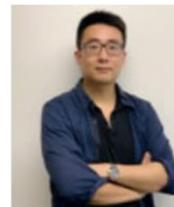

**Xi Zheng** (Member, IEEE) received the Ph.D. in Software Engineering from UT Austin, now Director of Intelligent systems research center (itseg.org) at Macquarie University. He specialized in Service Computing, IoT Security, and Reliability Analysis. Published more than 80 high-quality publications in top journals and conferences (PerCOM, ICSE, IEEE Communications Surveys and Tutorials, IEEE Transactions on Cybernetics, IEEE Transactions on Industrial Informatics, IEEE Transactions on Vehicular Technology, IEEE IoT journal, ACM Transactions on Embedded Computing Systems) and awarded multiple best papers in leading peer-reviewed international conferences. Guest Editor and PC members for top journals and conferences (IEEE Transactions on Industry Informatics, Future Generation Computer Systems, PerCOM, TrustCOM). WIP Chair for PerCOM 2020 and Track Chair for CloudCOM 2019. Publication Chair for ACSW 2019 and reviewers for many Trans journals and CCF A/CORE A* conferences.